\documentclass{article}
\usepackage{graphicx} %
\usepackage{notation}
\usepackage{natbib}
\usepackage{utopia}
\usepackage{cleveref}
\usepackage[dvipsnames]{xcolor}

\crefname{figure}{Fig.}{Figs.}
\crefname{algorithm}{Alg.}{Algs.}
\crefname{definition}{Def.}{Defs.}
\crefname{corollary}{Cor.}{Cors.}
\crefname{proposition}{Prop.}{Prop.}
\crefname{theorem}{Theorem}{Theorems}
\crefname{remark}{Remark}{Remarks}
\crefname{principle}{Principle}{Principles}
\crefname{lemma}{Lemma}{Lemmata}
\crefname{claim}{Claim}{Claims}
\crefname{table}{Tab.}{Tabs.}
\crefname{section}{Section}{Sections}
\crefname{subsection}{Section}{Sections}
\crefname{subsubsection}{Section}{Sections}
\crefname{assumption}{Assumption}{Assumptions}
\crefname{appendix}{App.}{App.}
\crefname{equation}{}{}
\crefname{example}{Example}{Examples}

\title{A Critical Perspective on Finite Sample Conformal Prediction Theory \\ in Medical Applications}
\author{Klaus-Rudolf Kladny \and Bernhard Schölkopf \and Lisa Koch \and Christian F. Baumgartner \and Michael Muehlebach}

\author{
  Klaus-Rudolf Kladny\textsuperscript{1, 2} \quad
  Bernhard Schölkopf\textsuperscript{1, 2, 3} \quad
  Lisa Koch\textsuperscript{4, 5, 6} \\
  Christian F. Baumgartner\textsuperscript{7, 8} \quad
  Michael Muehlebach\textsuperscript{1}
}

\date{
\textsuperscript{1} Max Planck Institute for Intelligent Systems, Germany \\
\textsuperscript{2} Tübingen AI Center, Germany \\
\textsuperscript{3} ELLIS Institute Tübingen, Germany \\
\textsuperscript{4} Department of Digital Medicine, University of Bern, Switzerland \\
\textsuperscript{5} Department of Diabetes, Endocrinology, Nutritional Medicine and Metabolism UDEM, Inselspital, Bern University Hospital, University of Bern, Switzerland \\
\textsuperscript{6} Diabetes Center Berne, Bern, Switzerland \\
\textsuperscript{7} University of Tübingen, Germany \\
\textsuperscript{8} University of Lucerne, Switzerland \\
}

\begin{document}

\maketitle

\begin{abstract}
    Machine learning (ML) is transforming healthcare, but safe clinical decisions demand reliable uncertainty estimates that standard ML models fail to provide. Conformal prediction (CP) is a popular tool that allows users to turn heuristic uncertainty estimates into uncertainty estimates with statistical guarantees. CP works by converting predictions of a ML model, together with a calibration sample, into prediction sets that are guaranteed to contain the true label with any desired probability. An often cited advantage is that CP theory holds for calibration samples of arbitrary size, suggesting that uncertainty estimates with practically meaningful statistical guarantees can be achieved even if only small calibration sets are available. We question this promise by showing that, although the statistical guarantees hold for calibration sets of arbitrary size, the practical utility of these guarantees does highly depend on the size of the calibration set. This observation is relevant in medical domains because data is often scarce and obtaining large calibration sets is therefore infeasible. We corroborate our critique in an empirical demonstration on a medical image classification task.
\end{abstract}

\section{Introduction}

Clinical decision-making demands trustworthy uncertainty estimates and factually grounded outputs. Although machine learning (ML) has delivered promising results across a range of medical applications—from breast cancer screening~\citep{laang2023artificial} to cardiovascular disease risk prediction~\citep{lubeck2025adaptable}—models remain prone to poor calibration, where stated uncertainties fail to reflect the true probability of being correct~\citep{van2019calibration}, and to hallucination, where outputs are not supported by facts or evidence~\citep{kim2025medical}. Therefore, deploying these systems in clinical decision processes can be dangerous, as these issues translate directly into serious physical consequences for patients if not addressed~\citep{asgari2025framework}. A concrete example is offered by~\citep{bromism2025chatgpt}, which describes a case in which a man developed bromism after consulting a large language model for dietary advice. To mitigate such risks, medical ML systems are considered medical devices and are subject to regulatory oversight to ensure their safety and effectiveness. Software as a medical device (SaMD) are for example regulated by the Food and Drug Administration (FDA) in the United States, or by the Medical Device Regulation (MDR) in Europe.

In both regions, regulators rely on standards and policy guidance which stress the importance of continuous monitoring, transparency, and the ability to interpret model outputs, particularly when models are adaptive or updated in deployment~\citep{amann2020explainability}. 
One key aspect of safety and regulatory compliance is the ability to quantify uncertainty in a reliable and interpretable manner, so that clinicians can assess the confidence of model outputs and make informed decisions, thereby ensuring that erroneous or unsafe model outputs can be detected and mitigated. Along these lines, the recently released consensus guideline for trustworthy and deployable artificial intelligence in healthcare (FUTURE-AI) explicitly demands that ML models provide calibrated uncertainty outputs as part of a ML system's traceability requirements.
In practice, calibrated uncertainty outputs can be integrated into clinical workflows in various ways. For example, they could support risk-based decision making, where thresholds can for example be applied to the uncertainty outputs for immediate action, ordering further tests, or monitoring only. Calibrated uncertainty outputs are also a useful communication tool to support shared decision-making between treating physicians and patients. Furthermore, information about model uncertainty can be useful for flagging or prioritizing cases for human review.

Conformal prediction (CP;~\citep{vovk2005algorithmic, shafer2008tutorial}) has emerged as a promising statistical framework to address this need. By means of a calibration dataset, CP transforms model predictions into prediction sets (multiple predictions, as shown in~\cref{fig:CP_example}), thereby providing a quantified measure of uncertainty. Under mild assumptions, these prediction sets come with a guarantee~\citep{Angelopoulos2023conformal}: for any new patient case, the set contains the true label with probability at least a user-specified level, independent of the model or task and, in theory, even for small calibration sets. CP has been applied successfully to a range of tasks, from image classification in histopathology~\citep{wieslander2020deep, olsson2022estimating} and dermatology~\citep{lu2022fair}, to quantile regression for retinal vessel segmentation~\citep{wundram2024conformal}, and natural language generation of radiology reports~\citep{kladny2025conformal}.

Various prior works identify limitations of CP regarding feature-conditional guarantees~\citep{vovk2012conditional, mehrtens2023pitfalls, chakraborti2025personalized}, non-exchangeability and distribution shift~\citep{barber2023conformal, mehrtens2023pitfalls}. While these concerns are theoretically sound and practically relevant, we argue that a fundamental mismatch between CP theory and practice remains: The assumption underlying guarantees that are invariant with respect to size of the calibration set. In particular, we show that an often-cited theoretical argument effectively presumes frequent recalibration on fresh calibration sets, which we deem infeasible in clinical practice. In addition, while calibration-set-conditional guarantees exist~\citep{vovk2012conditional}, these guarantees become practically meaningful only for very large calibration set sizes that may be costly or unattainable in clinical practice. We show on a histological image-classification dataset (\cref{sec:empirical_demonstration}) that coverage (the fraction of times the calibration sets contain the true label) conditional on a fixed, small calibration set can fall well below the desired coverage level with high probability. Consequently, in a realistic clinical workflow where calibration occurs only once (or infrequently), uncritical reliance on classical CP arguments can create an unjustified sense of safety and, at worst, lead to patient harm.

We proceed with a high-level introduction to the concrete practical method of CP in~\cref{sec:method}, followed by two theoretical arguments closely associated with the method in~\cref{sec:theory}. We then proceed to describe a practical workflow in~\cref{sec:practical_workflow} and how it mismatches the assumptions underlying the often-cited calibration-set-unconditional theory in~\cref{sec:mismatch}, followed by a real-world example on histological image classification in~\cref{sec:empirical_demonstration}. We discuss the clinical danger of over-reliance on CP theory in~\cref{sec:clinical_danger}, provide an outlook in~\cref{sec:outlook}, and conclude in~\cref{sec:conclusion}.

\section{Conformal Prediction Method} \label{sec:method}

While multiple variants of CP exist, we focus on a variant of CP called \textit{split conformal prediction}~\citep{papadopoulos2002inductive, lei2015conformal}. This variant has become the spotlight of attention in recent years, because it only requires training a ML model once. This aspect is relevant in the modern regime of training large neural network models, which are both, highly time and energy consuming~(e.g.,~\citep{strubell2020energy}).

We furthermore note that there exist various related methods to CP such as learn-then-test~\citep{angelopoulos2025learn}, PAC confidence sets~\citep{park2019pac} and risk-controlling prediction sets~\citep{bates2021distribution}. We stress that for these methods, the arguments made in the present work do not generally hold.

\begin{figure}[t]
    \centering
    \includegraphics[width=1\textwidth]{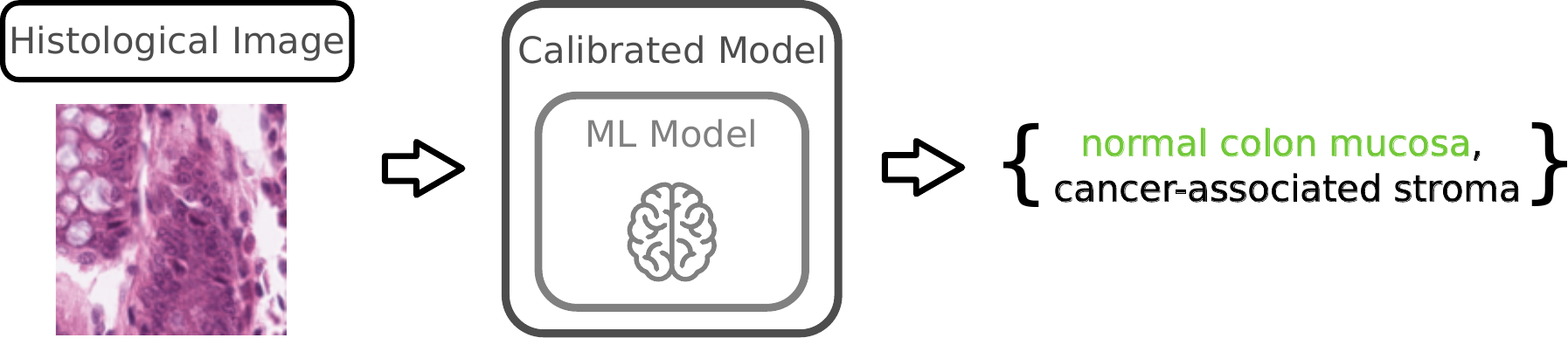}
    \caption{\textbf{CP for Histological Image Classification.} CP uses a input (in the example, a histological image) together with a calibration set (not visualized) to generate a prediction set, i.e., multiple labels. In the demonstrated example, the prediction set contains two labels, the correct label ``normal colon mucosa'' (green) and an incorrect one ``cancer-associated stroma'' (black).}
    \label{fig:CP_example}
\end{figure}

The split CP method (visualized in~\cref{fig:main_visualization}) splits a labeled data set $\Dcal = \{ (X_i, Y_i) \}_{i=1}^n$ with features $X_i$ and labels $Y_i$ into two different data sets: (1) A training set, denoted by $\Dtrain$; and (2) a calibration set, which we denote by $\Dcalib$. The training set $\Dtrain$ is used to train a ML model $\Mcal$, which creates probability estimates for different values of $Y$, given features $X$. After training, the model $\Mcal$ is calibrated using the calibration set $\Dcalib$ to a create prediction set $\Ccal_{\Mcal}(X; \Dcalib)$. For example, as shown in~\cref{fig:CP_example}, $X$ could correspond to a histological image and $\Ccal_{\Mcal}(X; \Dcalib)$ could contain the classes \textit{normal colon mucosa} and \textit{cancer-associated stroma} among a larger set of tissue classes. The size of the generated prediction set reflects the model uncertainty: If a prediction set contains many classes, the model is uncertain about the true class. If the prediction set contains few classes, the model is more certain about the true class.

For details about how conformal prediction sets are generated, we refer to~\citep{angelopoulos2021gentle}.

\section{Conformal Prediction Theory} \label{sec:theory}

In this section, we demonstrate two guarantees associated with the conformal prediction method: The calibration-set-unconditional theory~(\cref{sec:unconditional_theory}) is the most well-known one and is often referenced to legitimate the conformal prediction method~(\cref{sec:method}). In the present work, we question the clinical relevance of this guarantee and instead highlight a lesser-known calibration-set-conditional guarantee that remedies a core issue in the former guarantee~(\cref{sec:conditional_theory}). We will then discuss a practical workflow in~\cref{sec:practical_workflow} and how both guarantees can fail to be practically meaningful.

\subsection{Calibration-Set-Unconditional Theory} \label{sec:unconditional_theory}

The CP method~(\cref{sec:method}) is typically motivated by the guarantee that the true label $Y$ for a new, unseen case $X$ (that is, it is not included in the training or calibration set), is included in the prediction set $\Ccal_{\Mcal}(X; \Dcalib)$, with high probability. Specifically, for a user-defined parameter $\alpha \in (0, 1)$, the guarantee can be written as
\begin{equation} \label{eq:conformal_guarantee}
    \PP_{Y, X, \Dcalib} (Y \in \Ccal_{\Mcal}(X; \, \Dcalib)) \; \geq \; 1 - \alpha,
\end{equation}
where we refer to $1 - \alpha$ as the \textit{coverage level}. For instance, if $\alpha = 0.1$, then~\cref{eq:conformal_guarantee} tells us that the correct class $Y$ will be contained in $\Ccal_{\Mcal}(X; \, \Dcalib)$ with probability at least $90 \%$, i.e., we are guaranteed to achieve coverage at level $0.9$. Notably, this guarantee holds irrespective of how well the underlying ML model $\Mcal$ performs and how large the calibration set $\Dcalib$ is.

The key caveat is that is marginal and not conditional over the calibration set. Hence for an ``unlucky'' draw the prediction set $\Ccal_{\Mcal}(X; \, \Dcalib)$ may lead to poor coverage.

\subsection{Calibration-Set-Conditional Theory} \label{sec:conditional_theory}

An additional guarantee that is conditional on the calibration set size has been derived by~\citep{vovk2012conditional}. Specifically, defining $\tilde{\alpha} = \alpha + \epsilon$, for any $\epsilon > 0$, it can be shown that
\begin{equation*}
    \PP_{\Dcalib} ( \PP_{Y, X} (Y \in \Ccal_{\Mcal}(X; \, \Dcalib) \, | \, \Dcalib) \; \geq \; 1 - \tilde{\alpha}) \; \geq \; 1 - \delta 
\end{equation*}
for
\begin{equation*}
    \delta \, \geq \, \text{Binomial}_{m, \tilde{\alpha}} \left( \lfloor \alpha(m + 1) - 1 \rfloor \right),
\end{equation*}
where $\text{Binomial}_{m, \tilde{\alpha}}$ is the binomial cumulative distribution function with $m$ trials and probability of success $\tilde{\alpha}$.

\section{Practical Workflow} \label{sec:practical_workflow}

We now proceed to describe a concrete workflow of how conformal prediction is practical in a clinical setting.

\begin{figure}[t]
    \centering
    \includegraphics[width=1\textwidth]{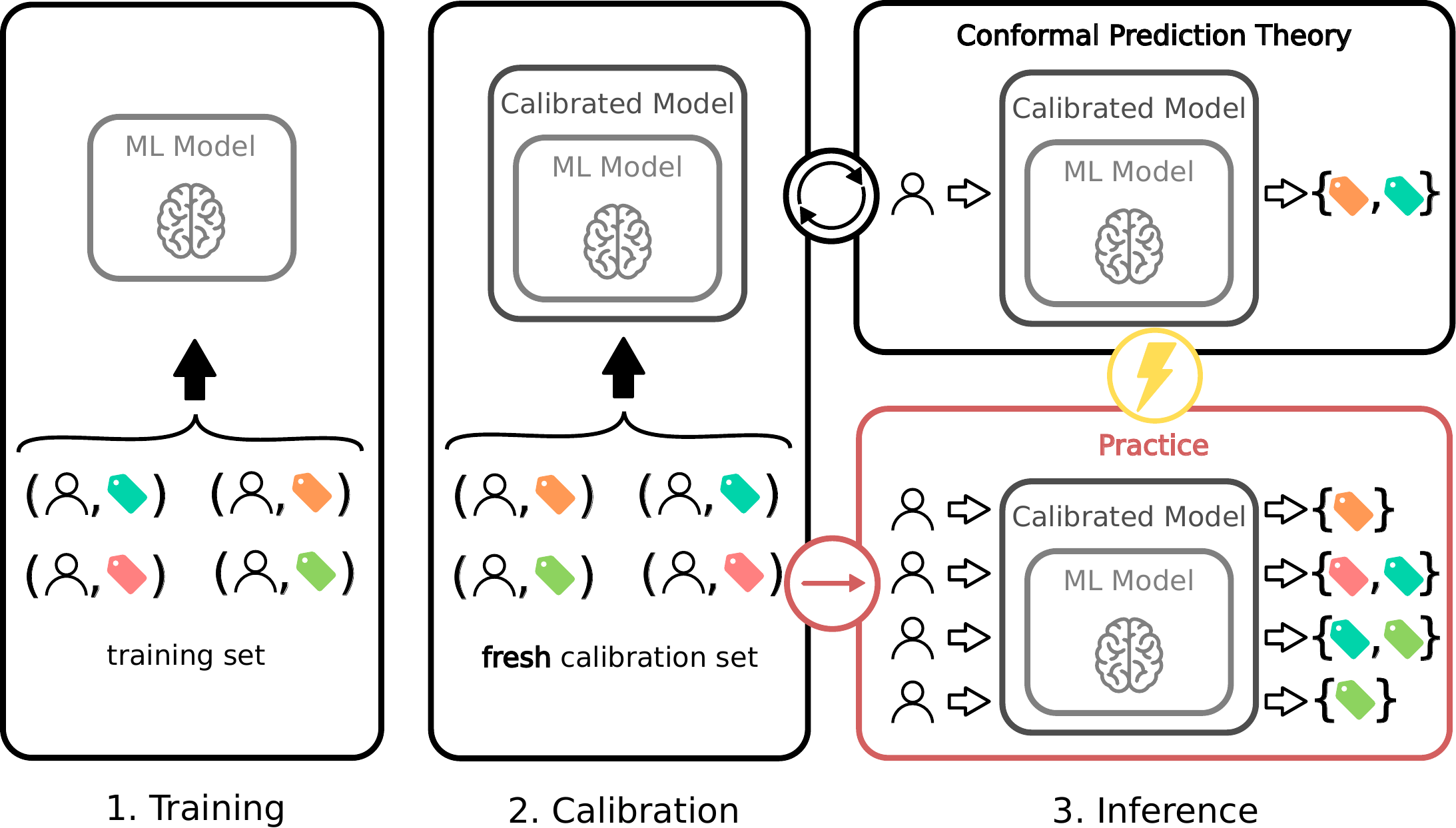}
    \caption{\textbf{Mismatch Between Theory and Practice.} The conformal prediction workflow can be split into three stages: training, calibration and inference. The standard calibration-set-unconditional guarantee~\cref{sec:unconditional_theory} is \textbf{marginal} over the calibration set: it presumes the model is recalibrated with a fresh set before each round of inference (top-right panel). The more practically feasible approach is to perform calibration once~(see \cref{sec:practical_workflow}), after which the same calibrated model is applied to many inference cases (bottom-right panel). In this single-calibration regime, what matters is the coverage \textbf{conditional} on the calibration set—yet CP theory provides either no conditional guarantees~(\cref{sec:unconditional_theory}) or conditional guarantees~(\cref{sec:conditional_theory}) that are expressive only for very large calibration sets.}
    \label{fig:main_visualization}
\end{figure}

To begin with, we assume that a reliably labeled data set has been created by domain experts and arbitrarily split into training set and calibration set, according to the CP method described in~\cref{sec:method}. The practical workflow is to perform the following three steps (visualized in the bottom variant of~\cref{fig:main_visualization}):  
\begin{enumerate}
    \item Train a ML model on the training data.
    \item Calibrate the trained model using a (potentially small) calibration set.
    \item Use the calibrated model to perform inference on new patients, \textbf{without (or with infrequent) re-calibration}.
\end{enumerate}
We see that steps (1) and (2) follow immediately from the methodological description of CP in~\cref{sec:method}. However, the CP method only describes how to generate prediction sets and does not make explicit statements about how to perform inference on multiple new patients. 

A natural presumption to make is that according to unconditional CP theory~(\cref{sec:unconditional_theory}), the desired coverage level $1 - \alpha$ will be attained after performing one calibration, i.e., we generate prediction sets for multiple new patients without re-calibrating the model regularly with a new calibration set (step (3)). However, we will demonstrate in the following section that this presumption is incorrect.

\section{Mismatch Between Theory and Practice \\ for Calibration-Set-Unconditional Guarantees} \label{sec:mismatch}

In this section, we elaborate on the theory underlying unconditional conformal prediction~(\cref{sec:unconditional_theory}), what the theory practically means and how it clashes with the setup sketched in~\cref{sec:practical_workflow}. Thereafter, we empirically demonstrate this point on a histological image classification task.

If we re-sample $Y$, $X$ and $\Dcalib$ many times to re-evaluate coverage, unconditional CP theory~(\cref{sec:unconditional_theory}) can be translated to the practical statement that the mean coverage will tend to $1 - \alpha$, irrespective of the size of $\Dcalib$. Formally, this means that 
\begin{equation} \label{eq:conformal_theory_argument}
    \frac{1}{M \cdot K} \sum_{j=1}^M \sum_{i=1}^K \indfonearg{y^{(i)} \in \Ccal_\Mcal(X^{(i)}; \Dcalib^{(j)})} \; \approx \; 1 - \alpha,
\end{equation}
where $y^{(i)}, X^{(i)}$ for $i = 1, 2, ..., K$ are independent realizations of labels and features and $\Dcalib^{(j)}$ for $j = 1, 2, ..., M$ are independent calibration sets, respectively. We note that for~\cref{eq:conformal_theory_argument} to hold, both $K$ and $M$ need to be large. However, we see that~\cref{eq:conformal_theory_argument} is not aligned with the workflow described in~\cref{sec:practical_workflow}: In the workflow from~\cref{sec:practical_workflow}, calibration of the model is performed \textbf{once}, instead of re-calibrating the model repeatedly with an entirely fresh calibration set. We argue that what matters for the workflow of~\cref{sec:practical_workflow} is that \textbf{conditionally on a single calibration set}, we obtain approximate $1 - \alpha$ coverage. This means that we instead strive for
\begin{equation} \label{eq:practical_workflow_desideratum}
    \frac{1}{K} \sum_{i=1}^K \indfonearg{y^{(i)} \in \Ccal_\Mcal(X^{(i)}; \Dcalib)} \; \approx \; 1 - \alpha, 
\end{equation}
for a single realization of the calibration set $\Dcalib$. However, the often-cited unconditional CP guarantee~(\cref{sec:unconditional_theory}) has no implications for the calibration-data-conditional statement~\cref{eq:practical_workflow_desideratum}. In fact, we empirically demonstrate in the following section that the achieved coverage can be far below the desired coverage level $1 - \alpha$, with high probability, if the calibration set is small. 

\begin{figure}[t]
    \centering
    \includegraphics[width=1\textwidth]{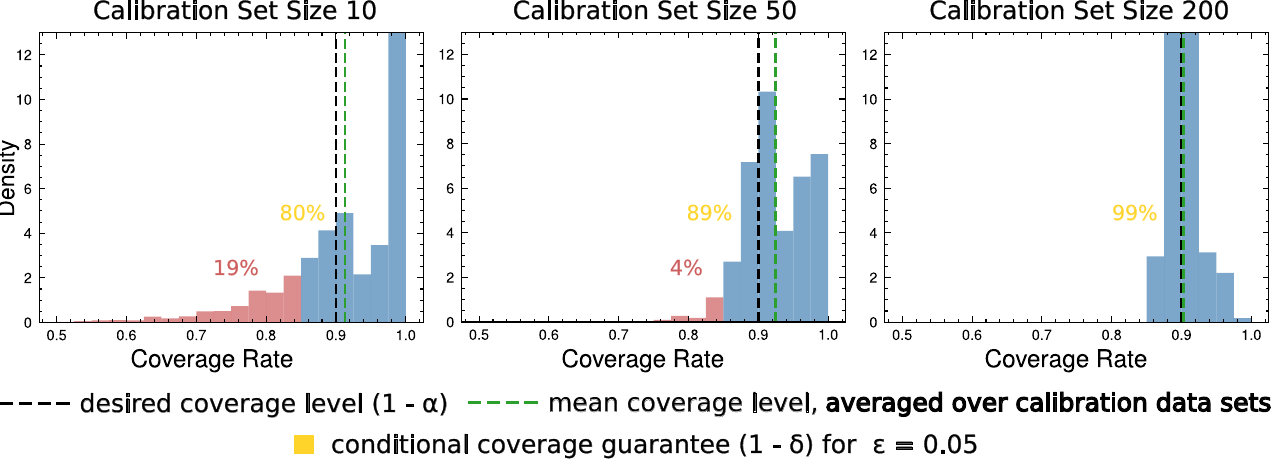}
    \caption{\textbf{Calibration-set–conditional coverage for histological image classification.} Each histogram shows the empirical distribution of conformal-prediction coverage over independent calibration sets of size $m=10$, $50$, and $200$. The vertical green line marks the mean unconditional coverage, which theory guarantees to exceed the nominal level $1-\alpha = 90\%$. Practical reliability, however, is determined by the spread: with only $m=10$ (left panel) almost one-fifth of calibration sets ($19\%$) deliver less than $85\%$ calibration-set-conditional coverage, whereas this shortfall only disappears once $m$ reaches $200$ (right panel). This circumstance is only taken into account by the (less well-known) calibration set conditional guarantee~(yellow;~\cref{sec:conditional_theory}), which yields practically useful guarantees only for very large calibration sets.}
    \label{fig:experimental_results}
\end{figure}

\subsection{Empirical Demonstration on Histological Image Classification} \label{sec:empirical_demonstration}

We consider a histological image classification task on the NCT-CRC-HE-100K dataset~\citep{kather2019predicting, yang2023medmnist}, where the goal is to classify nine tissue types from non-overlapping patches extracted from Hematoxylin and Eosin–stained histological images of colon tissue. The labels correspond to the tissue type visible in each patch, with two types being associated with colorectal cancer.

We use $10,000$ examples for training and split the remaining data into two data sets: The first split is used to calibrate the model, but we do not calibrate using the entire split. Instead, we further chunk this data set into sub-splits ranging from sizes $10$ to $200$, which we use for calibration. We then use the second split to assess the calibration-set-conditional coverage for individual calibration sets. 

The result, demonstrated in~\cref{fig:experimental_results}, shows histograms of the calibration set conditional coverage obtained by using calibration sets for three different sizes $m \in \{10, 50, 200\}$. The empirical experiment confirms CP theory~\cref{sec:theory}, which states that averaged over many different calibration sets, the coverage (green dashed line) is larger than the desired coverage level $90 \%$ (black dashed line). However, this guarantee has no implications for the spread of the distribution: For small calibration sets, $19\%$ of calibrations fall far below the desired coverage (below $85\%$), as can be seen be the red area of the histogram. Thus, if we do not regularly (and frequently) re-calibrate the model, the risk of achieving poor coverage is still very high. The calibration-set-conditional spread around the desired coverage level can only be decreased by choosing a larger calibration set, as can be seen in the right-most histogram of~\cref{fig:experimental_results}: The probability mass becomes more centered around the desired coverage level. While the calibration-set-conditional guarantee~(\cref{sec:conditional_theory}) does take calibration-set-conditional properties into account, we also see that guarantee is only expressive for large data sets.

\section{Clinical Danger of Relying on CP Theory} \label{sec:clinical_danger}

The greatest danger of the mismatch described in~\cref{sec:mismatch} lies in the fact that the calibration-set-unconditional theory~(\cref{sec:unconditional_theory}) may suggest that the size of the calibration set is irrelevant, because the theory holds true irrespective of the calibration set size. If, however, conformal prediction is used according to the setup described in~\cref{sec:practical_workflow}, the size of the calibration set is decisive for achieving coverage close to the desired level. Blind reliance on the classical CP argument~(\cref{sec:unconditional_theory}) can therefore foster an unwarranted sense of safety and, at worst, contribute to misdiagnosis. Such miscalibration is not merely a technical concern: in clinical contexts, it can translate into delayed or incorrect treatment decisions, potentially leading to severe consequences for patients. Moreover, it hampers the responsible deployment of machine learning systems in healthcare workflows - an area that would otherwise hold considerable promise for improving diagnostic accuracy and efficiency. Ultimately, repeated failures arising from misplaced trust in theoretical guarantees risk undermining clinicians’ and the public’s confidence in ML-assisted medical technologies. 

\section{Outlook}
\label{sec:outlook}
Looking ahead, we believe that advancing conformal prediction in medicine requires not only improving sample efficiency, but also fostering a shared understanding of what its statistical guarantees practically entail. Even perfectly valid mathematical guarantees can be misinterpreted when their operational meaning is not clearly communicated to clinicians, regulators, and other non-specialist stakeholders. In particular, while unconditional coverage guarantees may sound reassuring, their dependence on repeated recalibration or large calibration sets may easily be overlooked in practice. Bridging this gap demands both methodological and translational efforts: methodologically, by developing techniques that offer meaningful guarantees under realistic data limitations; and translationally, by creating communication standards and reporting practices that make explicit what can and cannot be expected from a deployed conformal predictor. In safety-critical domains like healthcare, such clarity is a prerequisite for trustworthy adoption.

\section{Conclusion} \label{sec:conclusion}

Conformal prediction is often promoted for its finite-sample coverage guarantees. Our empirical findings show that, while this claim is mathematically valid, its practical relevance relies highly on the concrete sample size. In fact, we deem that the often cited calibration-set-size-invariant guarantee~(\cref{sec:unconditional_theory}) may encourage inexperienced clinicians to rely on undersized calibration sets -- an oversight that could carry serious clinical consequences. Conformal prediction remains valuable, when sufficiently large calibration sets are available, and under consideration of (practically more relevant) calibration-set-conditional guarantees~(\cref{sec:conditional_theory}). By demonstrating the clinical importance of this issue, we hope to steer the community toward focusing on calibration-set-size-conditional conformal uncertainty quantification for small sample sizes.

\section{Competing Interests}

There exist no competing interests.

\bibliography{main}
\bibliographystyle{abbrv}

\end{document}